\documentclass[10pt,twocolumn,letterpaper]{article}

\usepackage{iccv}              
\usepackage{svg}

\usepackage[most]{tcolorbox}
\usepackage[T1]{fontenc}
\tcbset{
  promptbox/.style={
    colback=blue!5!white,
    colframe=blue!75!black,
    fonttitle=\bfseries,
    fontupper=\small,
    breakable,
    width=\linewidth,
    sharp corners,
    boxrule=0.8pt
  }
}

\definecolor{iccvblue}{rgb}{0.21,0.49,0.74}
\usepackage[pagebackref,breaklinks,colorlinks,allcolors=iccvblue]{hyperref}

\def\paperID{*****} 
\def\confName{ICCV}
\def\confYear{2025}

\title{AQUAH: Automatic Quantification and Unified Agent in Hydrology}

\author{
Songkun Yan$^{1}$ \quad Zhi Li$^{2}$ \quad Siyu Zhu$^{1}$ \quad Yixin Wen$^{3}$ \\ 
Mofan Zhang$^{4}$ \quad
Mengye Chen$^{1}$ \quad Jie Cao$^{5}$ \quad Yang Hong$^{1, *}$ \\[4pt]
$^{1}$School of Civil Engineering and Environmental Science, University of Oklahoma, Norman, OK, USA\\
$^{2}$Department of Earth System Science, Stanford University, Stanford, CA, USA\\
$^{3}$Department of Geography, University of Florida, Gainesville, FL, USA\\
$^{4}$Civil and Environmental Engineering, Stanford University, Stanford, CA, USA\\
$^{5}$School of Computer Science, University of Oklahoma, Norman, OK, USA\\[4pt]
{\tt\small skyan@ou.edu \quad li1995@stanford.edu \quad Siyu.Zhu-1@ou.edu \quad yixin.wen@ufl.edu}\\
{\tt\small mofanz@stanford.edu \quad mchen15@ou.edu \quad jie.cao@ou.edu \quad yanghong@ou.edu}
}

\begin{document}
\maketitle
\begin{abstract}
We introduce \textbf{AQUAH}, the first end-to-end language-based agent designed specifically for hydrologic modelling. Starting from a simple natural-language prompt (e.g., “simulate floods for the Little Bighorn basin from 2020 to 2022”), AQUAH autonomously retrieves the required terrain, forcing, and gauge data; configures a hydrologic model; runs the simulation; and generates a self-contained PDF report. The workflow is driven by vision-enabled large-language models, which interpret maps and rasters on the fly and steer key decisions such as outlet selection, parameter initialisation, and uncertainty commentary. Initial experiments across a range of U.S. basins show that AQUAH can complete cold-start simulations and produce analyst-ready documentation without manual intervention—results that hydrologists judge as clear, transparent, and physically plausible. While further calibration and validation are still needed for operational deployment, these early outcomes highlight the promise of LLM-centred, vision-grounded agents to streamline complex environmental modelling and lower the barrier between Earth-observation data, physics-based tools, and decision makers.
\end{abstract}    
\section{Introduction}
\label{sec:intro}

\paragraph{Motivation.} 
\vspace{-0.5em}

\noindent
Hydrologic simulation and Earth observation analysis are indispensable for managing water resources in a changing climate\citep{li2022conterminous,gao2021spatiotemporal,li2023decadal}. Yet fragmented workflows, steep technical requirements, and lengthy model‑setup times continue to restrict these capabilities—especially for non‑experts and rapid‑response applications\citep{carlton2016using,bahremand2016hess}. A genuinely transformative solution is therefore needed to dismantle these barriers and make advanced, equitable modeling tools broadly accessible.

\paragraph{Problem.} 
\vspace{-0.5em}

\noindent
Current hydrologic tools are not designed for accessibility or automation; setup and data processing are time-consuming \citep{mens2021dilemmas}. Users must often manually download data, configure models, and interpret outputs, requiring both domain knowledge and programming skills. Additionally, interpreting the results generated by the model is a barrier and requires years of related domain experience. Although this is not the sole problem for hydrologic science, we, in this paper, propose to bridge the gap and enhance communication of hydrologic simulation.

\paragraph{Solution.} 
\vspace{-0.5em}

\noindent
We present \textbf{AQUAH}---\textit{Automatic Quantification \& Unified Agent in Hydrology}---a next-generation, vision-enhanced large-language-model (LLM) agent that converts free-form user prompts into end-to-end hydrologic simulations and narrative reports. Riding on the rapid advances in vision multimodal LLMs~(VLMs), AQUAH leverages state-of-the-art vision capabilities to \emph{interpret maps, rasters, and other geospatial imagery on the fly}, replacing several expert-driven decisions---such as outlet selection, parameter initialization---with reliable, data-driven automation that has already shown promising accuracy and consistency in our experiments.

\medskip
\noindent
 Our agent AQUAH stitches together geospatial data retrieval, Earth-observation forcing data, hydrologic models (e.g., Coupled Routing and Excess Storage, \textsc{crest}~\citep{wang2011coupled}), and automated visualization in a seamless workflow. Thanks to its \emph{data-agnostic, model-agnostic, plug-and-play} design, AQUAH lowers the entry barrier for users without technical modeling backgrounds while still satisfying domain experts. By demonstrating how vision-enabled LLMs can assume formerly human-exclusive roles, AQUAH points the way toward fully autonomous hydrologic modeling agents.

\section{Related Work}
\label{sec:Related}

\paragraph{Multimodal and Tool-Augmented LLMs for Scientific Reasoning.}
Recent advances in LLMs have demonstrated remarkable capabilities in scientific reasoning when paired with external tools and multimodal inputs~\citep{sanderson2023gpt, liu2024deepseek, li2024survey}. Frameworks such as ReAct, Toolformer, and HuggingGPT combine language understanding with programmatic control, enabling agents to interface with APIs, code environments, and databases ~\citep{yao2023react,shen2023hugginggpt}. Emerging multimodal foundation models (e.g., GPT-4o~\citep{gpt4o}, Gemini~\citep{anthropic2025claudesonnet4}, Kosmos-2~\cite{peng2023kosmos2}) have shown promise in parsing text, images, and structured data for scientific workflows ~\citep{sanderson2023gpt,team2024gemma}. However, their application to Earth system science remains limited, especially for domain-specific physical modeling like hydrology. AQUAH builds on this foundation by integrating natural language processing with geospatial data tools, Earth observation inputs, and model execution capabilities.

\paragraph{Earth Observation + AI for Sustainability.}
The fusion of EO data and AI has advanced rapidly in applications like land cover classification, crop monitoring, and disaster mapping \citep{jakubik2025terramind, blumenstiel2025terramesh,yan2023pcssr}. Vision-based foundation models (e.g., Segment Anything Model, SatMAE, Prithvi) have pushed the frontier in remote sensing understanding \citep{szwarcman2024prithvi}. Yet, most works focus on static scene understanding rather than simulation-driven analysis. In hydrology, EO data like CHIRPS rainfall or MODIS evapotranspiration are used in modeling pipelines, but are rarely integrated via intelligent agents or prompted via natural language.

\paragraph{Automation in Hydrologic Modeling.}
Traditional hydrologic models (e.g., CREST, EF5, SWAT, HEC-HMS) are well-established for flood simulation and watershed analysis \citep{li2021crest, wang2011coupled, gassman2014applications}. However, they require significant manual effort for setup, data integration, parameter calibration, and output interpretation. Recent efforts in workflow automation (e.g., RavenPy, RavenWPS) have improved usability \citep{arsenault2023pavics}, but these tools are not conversational, nor are they driven by natural language or LLMs. Our work fills this gap by combining the rigor of physics-based models with the accessibility of LLM agents, forming a bridge between EO, simulation, and narrative reporting.

\section{AQUAH}
\label{sec:AQUAH}

\subsection{System Architecture Overview}

We design \textbf{AQUAH} as a modular language-agent framework that bridges natural language interaction with Earth observation data, geospatial processing, and hydrologic simulation tools. The architecture (Figure~\ref{fig:architecture}) consists of:
\begin{figure}[htbp]
  \centering
  \includegraphics[width=0.9\linewidth]{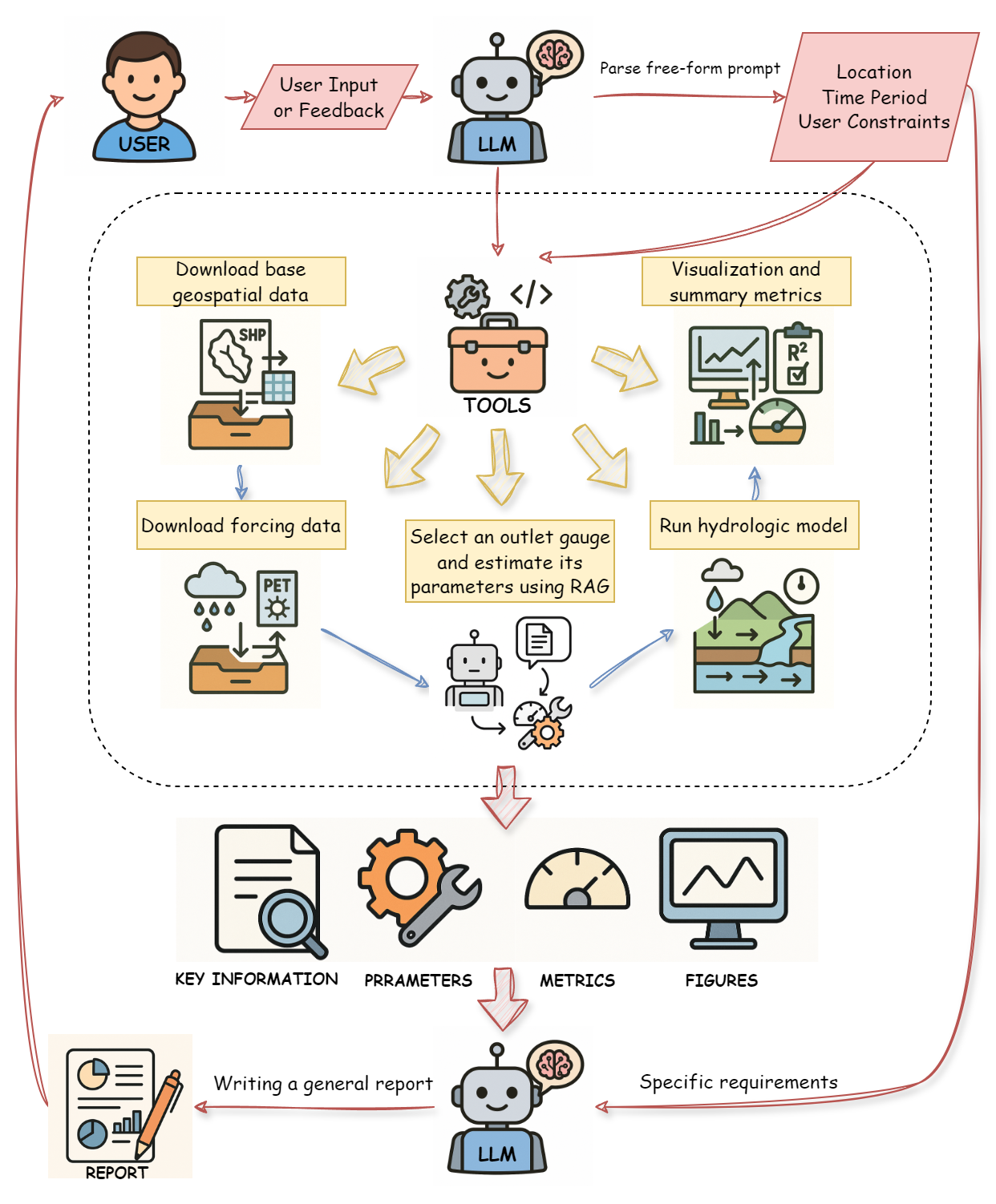}
  \caption{Overview of the AQUAH architecture showing key components and data flow.}
  \label{fig:architecture}
\end{figure}
\begin{itemize}
    \item \textbf{LLM Interface:} Converts user-provided natural language inputs into structured simulation instructions specifying locations, time periods, and analytical goals.
    \item \textbf{Tool Executor Layer:} Manages and executes Python-based geospatial libraries, hydrologic model wrappers, visualization routines, and statistical summarization tools, orchestrating comprehensive data retrieval, simulation, and analysis workflows. 
    \item \textbf{Dynamic Data Pipeline:} Automatically fetches essential hydrological data such as digital elevation models (DEM), precipitation, potential evapotranspiration (PET), and observed discharge datasets, based on user input.
    \item \textbf{Hydrologic Model Integration:} Implements the CREST model for hydrological simulations, utilizing dynamically obtained datasets and providing initial parameter estimates informed by Retrieval-Augmented Generation (RAG) and LLM reasoning.
    \item \textbf{Report Generation Engine:} Automatically compiles simulation outcomes, visualizations, and analytical summaries into structured, publication-quality Markdown or PDF reports.
    \item \textbf{Interactive Feedback Loop:} Allows users to refine simulations via natural-language feedback—e.g., selecting alternative gauges or adjusting parameters. The LLM parses these requests, updates model configurations through the Tool Executor Layer, reruns the hydrologic simulation, and regenerates an updated report, enabling rapid, iterative scenario exploration.

\end{itemize}

AQUAH supports fully automated hydrologic simulations driven entirely by natural language requests, leveraging LLM-powered image interpretation and decision-making capabilities. It operates within data-available regions, particularly across the contiguous United States (CONUS), enabling both researchers and non-technical users to effortlessly conduct detailed hydrologic analyses.

\subsection{Earth Observation}
\label{subsec:eo-datasets}
AQUAH automatically harvests the inputs required for hydrologic simulation: (i) basin outlines from the U.S.\ Geological Survey (USGS); (ii) terrain products—including a Digital Elevation Model, Drainage-Direction Map, and Flow-Accumulation Map—directly from the HydroSHEDS archive; (iii) precipitation forcing from the Multi-Radar/Multi-Sensor (MRMS) system and potential evapotranspiration (PET) fields from USGS, each clipped to the basin envelope; and (iv) in-situ discharge records served by USGS web APIs. Any missing files trigger fallback notifications and sensible default values, so the workflow remains robust across basins with heterogeneous data coverage. See Appendix~\ref{app:setup} for more information.

\subsection{Hydrologic Model}
\label{subsec:hydro-models}
For runoff generation and routing, AQUAH employs the distributed CREST~(Coupled Routing and Excess STorage) model \citep{wang2011coupled,flamig2020ensemble}. CREST solves basin water balance components—precipitation partitioning, infiltration, evapotranspiration, and subsurface exchange—and propagates the resulting flows using a kinematic-wave scheme. Model parameters are exposed for basin-specific calibration, with first-guess values supplied automatically by AQUAH’s language-agent modules. Details are in Appendix~\ref{app:crest}.

\subsection{Multi-Agent Architecture}
\begin{figure}[htbp]
  \centering
  \includegraphics[width=0.9\linewidth]{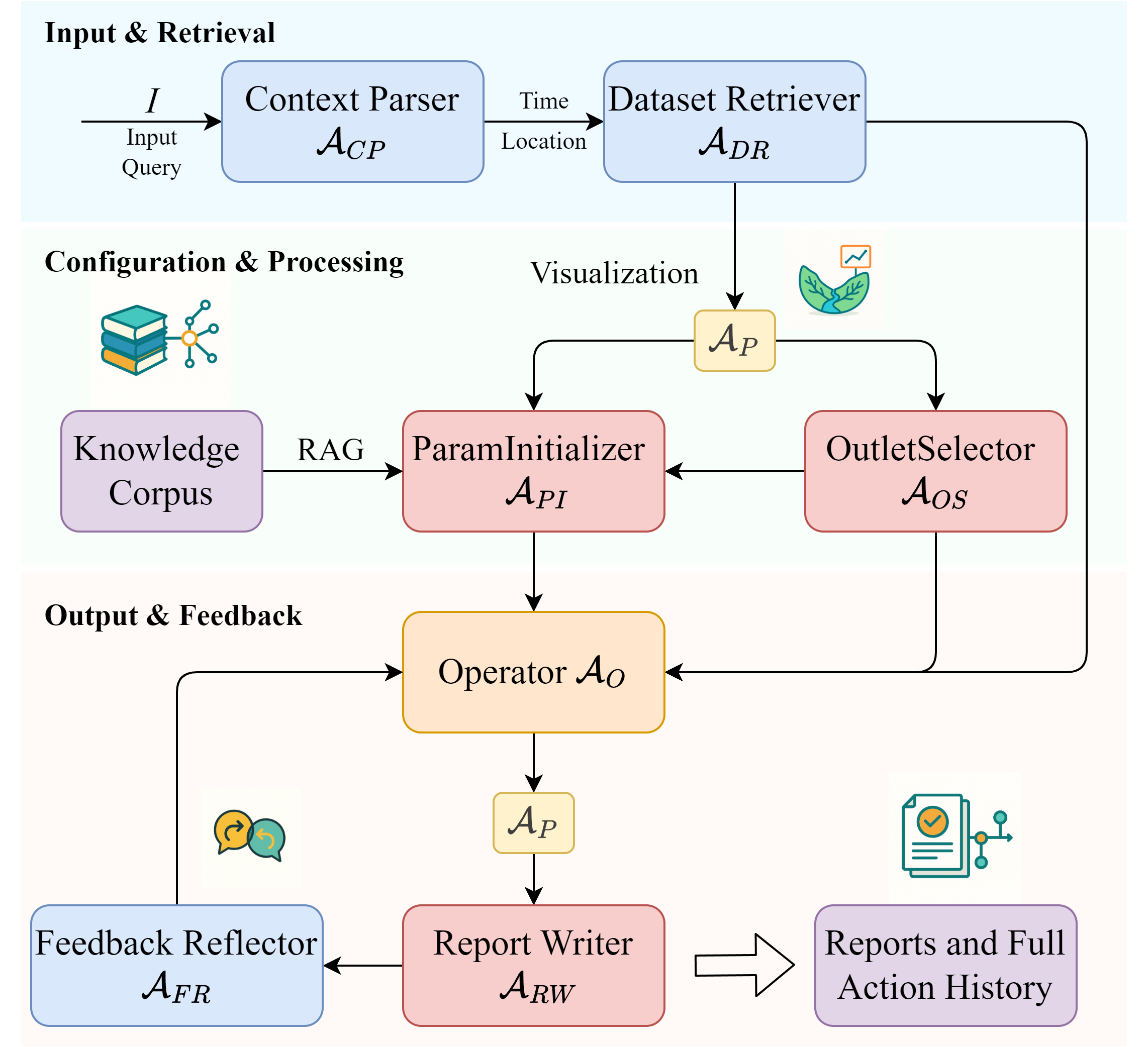}
    \caption{AQUAH multi-agent workflow across input–retrieval, configuration–processing, and output–feedback stages. Agents: \(\mathcal{A}_{\mathrm{CP}}\), \(\mathcal{A}_{\mathrm{DR}}\), \(\mathcal{A}_{\mathrm{P}}\) (Perceptor), \(\mathcal{A}_{\mathrm{OS}}\), \(\mathcal{A}_{\mathrm{PI}}\), \(\mathcal{A}_{\mathrm{O}}\), \(\mathcal{A}_{\mathrm{RW}}\), and \(\mathcal{A}_{\mathrm{FR}}\).}

  \label{fig:agent_connection}
\end{figure}

As shown in Figure~\ref{fig:agent_connection}, AQUAH is implemented as a \emph{multi-agent system}: a collection of specialized, communicating agents—denoted \(\mathcal{A}_{\!*}\)—that transform a free-form hydrologic modeling request into reproducible simulations, diagnostics, and reports.  Each agent owns a well-defined responsibility and passes structured artifacts to the next, enabling transparent reasoning, easier debugging, and seamless extensibility to additional Earth-system tasks.

\begin{itemize}
  \item \textbf{Context Parser Agent} (\(\mathcal{A}_{\mathrm{CP}}\)) – Parses the user’s natural-language request to extract structured simulation metadata—geographic extent, temporal window, and any special constraints—and forwards this normalized envelope to the Dataset Retriever~\(\mathcal{A}_{\mathrm{CP}}\), ensuring that all subsequent data acquisition is geographically and temporally aligned with the user’s intent.
  
  \item \textbf{Dataset Retriever Agent} (\(\mathcal{A}_{\mathrm{DR}}\)) – Receives the spatio-temporal envelope emitted by \(\mathcal{A}_{\mathrm{CP}}\); retrieves all mandatory forcing data and baseline geospatial layers (e.g., DEM, land cover, soil) intersecting that envelope; clips each layer to the provisional basin mask; generates quick-look visualisations that are forwarded to the Perceptor~\(\mathcal{A}_{\mathrm{P}}\) for morphological analysis feeding the OutletSelector~\(\mathcal{A}_{\mathrm{OS}}\) and ParamInitializer Agent~\(\mathcal{A}_{\mathrm{PI}}\); and converts the curated datasets~(mentioned in \S\ref{subsec:eo-datasets}) into the file formats and directory schema required by the Operator~\(\mathcal{A}_{\mathrm{O}}\).
  
  \item \textbf{Perceptor Agent}~(\(\mathcal{A}_{\mathrm{P}}\)) – Serves as the vision-perception module: ingests the visual artefacts rendered by \(\mathcal{A}_{\mathrm{DR}}\) (e.g., DEMs, flow-accumulation maps, preliminary hydrographs), employs a vision-augmented LLM to extract quantitative descriptors of basin morphology, drainage structure, and candidate gauge sites; supplies these descriptors to the OutletSelector Agent~\(\mathcal{A}_{\mathrm{OS}}\) and ParamInitializer Agent~\(\mathcal{A}_{\mathrm{PI}}\), and later interprets simulated versus observed hydrographs to deliver expert diagnostics for iterative refinement.
  
  \item \textbf{OutletSelector Agent} (\(\mathcal{A}_{\mathrm{OS}}\)) – Consumes the candidate gauge inventory and drainage descriptors extracted by \(\mathcal{A}_{\mathrm{P}}\); applies hydrologic heuristics encoded in the system prompt—such as favouring the gauge closest to the pour point, with long, gap-free records and minimal upstream regulation—to rank the options and designate the optimal basin outlet; emits the selected gauge’s identifier and coordinates to both the ParamInitializer~\(\mathcal{A}_{\mathrm{PI}}\) and Operator~\(\mathcal{A}_{\mathrm{O}}\) for subsequent simulation steps. The top example of Figure~\ref{fig:vision_agent} further illustrates the input and output of this gauge selection process. We also provide a detailed case study for VLM-based gauge selection in Section~\ref{subsec:case-study}.

  \item \textbf{ParamInitializer Agent} (\(\mathcal{A}_{\mathrm{PI}}\)) – Retrieves and parses domain manuals, peer-reviewed literature, and authoritative web resources via RAG to map each model parameter’s physical meaning and admissible range; ingests the processed basin attributes from \(\mathcal{A}_{\mathrm{P}}\), the forcing datasets prepared by \(\mathcal{A}_{\mathrm{DR}}\), and the selected outlet metadata; integrates all evidence to generate a basin-specific, physically plausible initial parameter vector that will seed the Operator Agent’s first simulation run.

  \item \textbf{Operator Agent} (\(\mathcal{A}_{\mathrm{O}}\)) – Ingests the forcing datasets~(\S\ref{subsec:eo-datasets}) and static rasters curated by~\(\mathcal{A}_{\mathrm{DR}}\), the outlet definition supplied by \(\mathcal{A}_{\mathrm{OS}}\), and the initial parameter vector crafted by \(\mathcal{A}_{\mathrm{PI}}\); configures the selected hydrologic models~(\S\ref{subsec:hydro-models}) with these inputs; executes the simulation over the user‐defined time window; captures full time-series outputs and performance diagnostics such as NSCE, RMSE, and bias; and packages these artefacts for downstream consumption by the Report Writer~\(\mathcal{A}_{\mathrm{RW}}\) and Feedback Reflector~\(\mathcal{A}_{\mathrm{FR}}\).

  \item \textbf{Report Writer Agent} (\(\mathcal{A}_{\mathrm{RW}}\)) – Consolidates the simulation outputs and diagnostics from \(\mathcal{A}_{\mathrm{O}}\) with the hydrograph analyses and visual artefacts supplied by \(\mathcal{A}_{\mathrm{P}}\); weaves in contextual metadata captured throughout the pipeline (basin description, forcing sources, parameter settings); and auto-compiles a structured, reader-friendly report enriched with maps, hydrographs, statistics, and explanatory narrative for delivery to the end user.

  \item \textbf{Feedback Reflector Agent} (\(\mathcal{A}_{\mathrm{FR}}\)) – Parses user commentary on the delivered report, updates the internal knowledge state, and, when revisions are warranted, transmits explicit re-run directives—updated parameters, alternative datasets, or extended periods—to the \(\mathcal{A}_{\mathrm{O}}\), thereby closing the human-in-the-loop calibration loop.

\end{itemize}

Based on the above agents, AQUAH autonomously orchestrates the full hydrologic modelling pipeline—parsing user requests, retrieving and preparing geospatial and forcing data, extracting morphological descriptors via vision-augmented LLMs, selecting optimal gauge outlets, initializing model parameters, running simulations, generating diagnostic reports, and incorporating user feedback—without requiring domain expertise or manual intervention.

\section{Design of VLM-based Agents}
\label{sec:vision_agents}

\begin{figure}[htbp]
  \centering
  \includegraphics[width=1\linewidth]{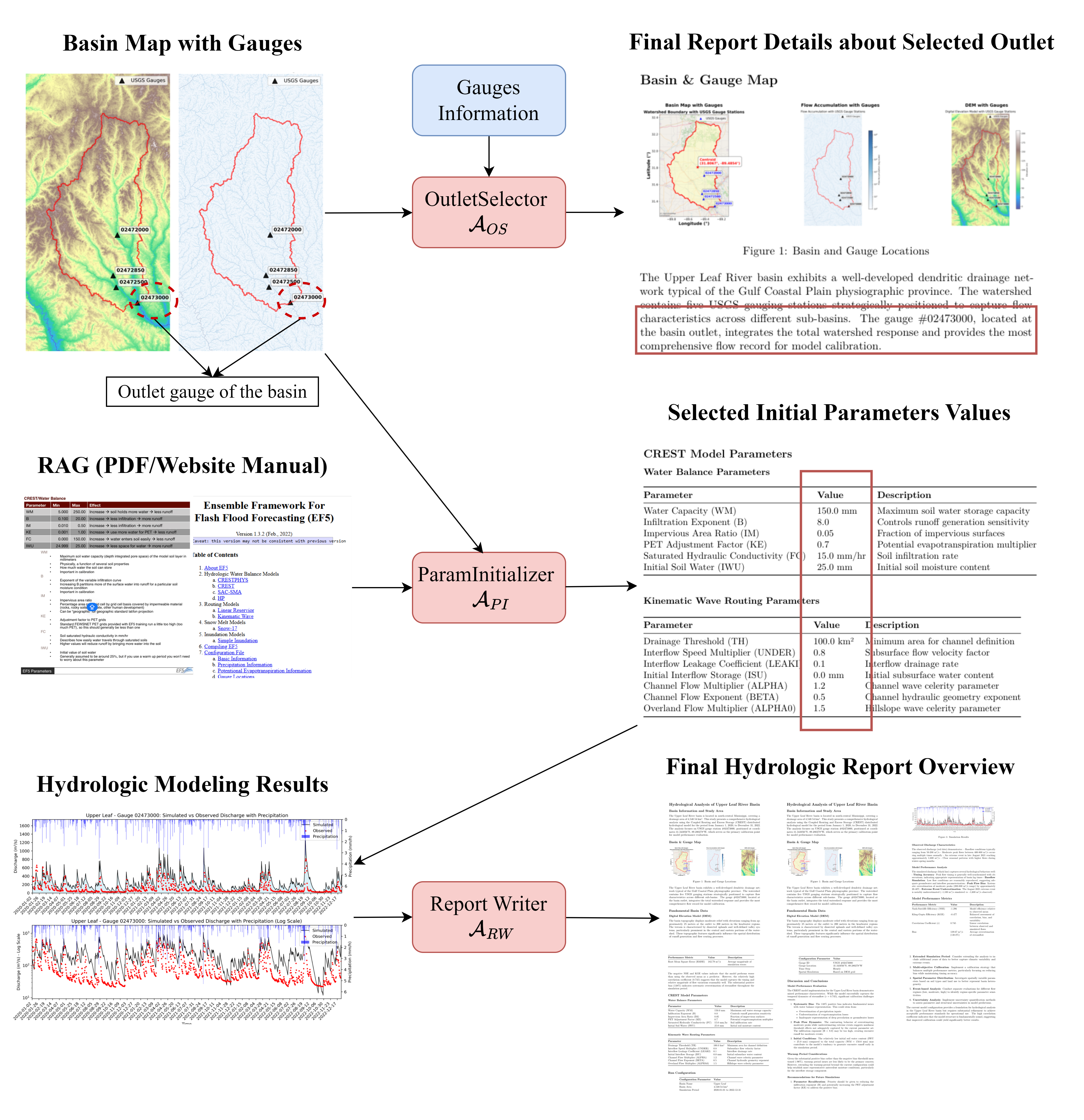}
    \caption{Vision–reasoning interplay within the \textsc{AQUAH} pipeline.  
The Perceptor Agent (\( \mathcal{A}_{\mathrm{P}} \)) transforms visual artefacts into structured cues that steer three downstream modules: (i) basin‐map rasters yield gauge locations that the OutletSelector (\( \mathcal{A}_{\mathrm{OS}} \)) ranks to pick the outlet; (ii) geomorphic attributes extracted from the same imagery combine with RAG-sourced documentation to let the ParamInitializer (\( \mathcal{A}_{\mathrm{PI}} \)) populate an initial parameter table; and (iii) simulated hydrographs are interpreted to provide narrative context for the Report Writer (\( \mathcal{A}_{\mathrm{RW}} \)). The panels on the right illustrate how each branch of the vision workflow materialises in the final deliverable: outlet information, auto-selected parameter values, and a consolidated hydrologic analysis overview.}

  \label{fig:vision_agent}
\end{figure}

AQUAH embeds cross-modal reasoning by equipping three key agents with large-language-model \emph{vision} capabilities:

\paragraph{\(\mathcal{A}_{\mathrm{OS}}\).}
Guided by the \emph{Outlet Gauge Selector} prompt, the vision–LLM pair first scans the basin map, DEM, and flow-accumulation layers to automatically list every \emph{candidate} station and its attributes.  
It then applies the ordered rules embedded in the prompt—exclude gauges on reservoirs, favor the lowest-elevation pour point with the largest drainage area and longest gap-free USGS record, and perform a final upstream-reservoir sanity check—iterating until a single outlet meets all criteria and hydrologic common sense.

After ingesting the basin-with-gauges image (Fig.~\ref{fig:vision_agent}, upper-left), the agent returns two plain-text lines:

\texttt{Selected gauge:\, [ID]} and \texttt{Explanation:\, [brief justification]}, providing both a machine-readable choice and a concise, human-readable rationale.

\begin{tcolorbox}[promptbox, title={Outlet Gauge Selector Prompt}]
You are a hydrologist who can interpret maps and select the most
appropriate USGS gauge to represent the **natural** basin outlet.
The user supplies: (1) a base-map with watershed boundary,
(2) a DEM with gauges, and (3) a flow-accumulation map with gauges.

Apply the following ordered rules when selecting \textbf{ONE} gauge
(earlier rules override later ones):\\
0) If the user's text clearly mentions or implies a specific gauge,
   city, or location, select it.\\
1) \emph{Exclusion} – disqualify any gauge located downstream of,
   or directly on, a reservoir/lake.\\
2) From the remaining gauges, prefer the one at the lowest-elevation
   point on the basin boundary where flow naturally exits (use DEM).\\
3) Prefer gauges capturing the largest drainage area and highest
   flow-accumulation values.\\
4) Prefer gauges with extensive, reliable USGS discharge records.\\
5) \emph{Second verification} – re-check that the chosen gauge is upstream
   of all reservoirs/lakes and sits at a natural outlet. If not,
   discard it and re-evaluate.

\textbf{Return} your response in this format:\\
\texttt{Selected gauge: [gauge ID number]}\\
\texttt{Explanation: [brief justification]}
\end{tcolorbox}

\paragraph{\(\mathcal{A}_{\mathrm{PI}}\).}
The initializer prompt is fed by two descriptive inputs.  
\emph{Basin description} (\texttt{\{basin\_desc\}}) is produced by a VLM that “reads’’ the basin-map images~(Fig.~\ref{fig:vision_agent}), then writes a short paragraph summarising key traits such as drainage area, relief, and dominant slope classes.  
\emph{Parameter guide} (\texttt{\{guide\}}) comes from a RAG pipeline that scans PDF manuals and web pages, condensing each source into plain-language hints on plausible CREST parameter ranges.  
Given these two narrative snippets, \(\mathcal{A}_{\mathrm{PI}}\) returns a one-line JSON object: a full CREST parameter vector plus a brief justification for every value, providing both machine-ready inputs and transparent reasoning.

\begin{tcolorbox}[promptbox, title={CREST Parameter Initializer Prompt}]
You are a hydrologist. Using the parameter guide and basin description,
propose first‐guess CREST parameters.

\textbf{Basin description (from LLM \emph{image interpretation})}:%
  \texttt{\{basin\_desc\}}\\
\textbf{Parameter guide (from LLM \emph{document/web summarisation})}:%
  \texttt{\{guide\}}

Return \underline{exactly} one line of JSON:\\
\texttt{\{"code":"crest\_args = types.SimpleNamespace(wm=<value>, b=<value>, im=<value>, ...",}\\
\texttt{"explanation":"each param justified in 100--300 words"\}}\\
No Markdown, no extra keys.
\end{tcolorbox}

\paragraph{\(\mathcal{A}_{\mathrm{RW}}\).}
Once the run finishes, \(\mathcal{A}_{\mathrm{RW}}\) fills the \emph{Hydrological Report Writer} prompt with two text fragments:  
\texttt{summary}—an auto-generated paragraph that turns stored run metadata (basin name, simulation window, chosen gauge, key metrics) into plain language—and  
\texttt{figures\_description}—sentences returned by the vision LLM after inspecting the maps and hydrographs.  
Guided by the prompt checklist, the agent assembles a Markdown file that includes (i) a title and basin locator map; (ii) cartographic layers for spatial context; (iii) rainfall-and-discharge plots; (iv) a table of CREST parameters and NSE, KGE, CC, bias, RMSE scores; and (v) a short discussion of results and recommended next steps.  
The Markdown is then rendered to PDF, giving users a compact, self-contained overview of basin features, model behaviour, and forecast quality (see report overview in Fig.~\ref{fig:vision_agent},lower right).


\begin{tcolorbox}[promptbox, title={Hydrological Report Writer Prompt}]
\textbf{Description}: Using the provided simulation metadata and results\\
\quad– \texttt{summary: \{summary\}}\\
\quad– \texttt{figures description: \{figures\_desc\}}\\[0.3em]
generate a complete Markdown report containing:

\begin{enumerate}[leftmargin=*]
  \item \textbf{Title and Basin Information}\\
        – Level-1 heading with basin name.\\
        – Basin \& gauge map, basic data and brief introduction.
  \item \textbf{Analysis Sections}\\
        – Simulation vs observation comparison.\\
        – Model performance metrics.\\
        – CREST parameters.\\
        – Conclusion/Discussion.
  \item \textbf{Required Images} (\texttt{![]()}):\\
        \texttt{combined\_maps.png}, \texttt{results.png}
  \item \textbf{Data Tables} (run arguments, metrics, parameters) – vertical listing.
  \item \textbf{Discussion Points}\\
        – Model performance evaluation.\\
        – Warm-up period considerations if bias $< -90\%$.\\
        – Recommendations for future runs.
\end{enumerate}

\textbf{Expected output}: a complete, publication-ready Markdown report
(no extra text after the report).
\end{tcolorbox}

\begin{table*}[t]
    \centering
    \begin{tabular}{lcccc|c}
        \hline
        \textbf{Model} & \textbf{Model Comp.} & \textbf{Sim. Results} & \textbf{Reasonableness} & \textbf{Clarity} & \textbf{Average} \\
        \hline
        \textbf{claude-4-opus}   & \textbf{7.51} & 5.60 & \textbf{6.97} & \textbf{7.95} & \textbf{7.01} \\
        claude-4-sonnet          & 7.43 & 5.46 & 6.77 & 7.49 & 6.79 \\
        gpt-4o                   & 6.74 & \textbf{5.89} & 6.06 & 6.51 & 6.30 \\
        o1                       & 7.11 & 4.80 & 6.23 & 6.94 & 6.27 \\
        gemini-2.5-flash         & 6.91 & 4.68 & 6.28 & 6.57 & 6.11 \\
        \hline
    \end{tabular}
    \caption{Quantitative evaluation of hydrological-report generation.
             Bold values mark the best score in each column.}
    \label{tab:hydro_eval}
\end{table*}

\section{Experiments}
\label{sec:experiment}

\subsection{Settings}

\paragraph{Hydrologic backbone and geospatial toolkit.}
AQUAH couples the distributed \textit{CREST} model with standard open-source GIS utilities (GDAL, Rasterio, Shapely, and Folium) for raster reprojection, vector clipping, and mapping.

\paragraph{Large-language models.}
Three vision-capable LLMs are benchmarked: GPT-4o (OpenAI) \citep{gpt4o}, Claude-Sonnet-4 (Anthropic) \citep{anthropic2025claudesonnet4}, and Gemini-2.5-Flash (Google) \citep{google2025gemini25flash}.  
Please refer to Appendix \ref{app:setup} for further details.

\paragraph{Earth-observation inputs.}
Daily precipitation forcing is provided by MRMS (1 km grid); potential evapotranspiration by USGS FEWS NET (1° grid); terrain layers—DEM, drainage-direction, and flow-accumulation—by HydroSHEDS at 3-arc-second~(~90 m) resolution; and discharge observations by USGS NWIS, which also supplies gauge metadata such as drainage area.

\paragraph{End-to-end automation.}
User prompts are decomposed by AQUAH’s agent stack into structured tasks that autonomously handle data download, model runs, and post-processing.  The workflow therefore spans geospatial preprocessing, Earth-observation ingestion, hydrologic simulation, and report generation without manual intervention.

\subsection{Benchmark \& Test Results}
\label{subsec:human_llm_eval}

To quantify the quality of the hydrological–simulation reports produced by our \emph{AQUAH}, we conducted a two–tier evaluation. All reports were anonymized and randomly ordered to ensure that evaluators were blind to the source model, thereby eliminating potential bias.

\paragraph{Domain–expert review.}
Several professional hydrologists were asked to
score each report on a 10-point scale (1 = poor, 10 = excellent) along four
facets that are critical for decision-grade hydrological studies:
\emph{Model Completeness}, \emph{Simulation Results}, \emph{Reasonableness}, and
\emph{Clarity}. Figure\ref{fig:report_review_example} in Appendix~\ref{app:eval} shows the detailed human judge interface.
\paragraph{LLM co-evaluation.}
In addition, we used the latest OpenAI \texttt{gpt-o3} model as an impartial,
large-scale language model (LLM) judge. This hybrid protocol mitigates individual-expert
variance while leveraging the consistency of an automated evaluator.

For each axis we take the arithmetic mean of the human and LLM scores; the
Average column is the unweighted mean across the four axes.  The numerical
results are summarised in Table~\ref{tab:hydro_eval}.  Overall, \texttt{claude-4-opus}
achieves the highest average score (7.01), leading or tying on three of the four
criteria, and outperforming all other contenders by at least 0.22 points. Refer to Appendix~\ref{app:eval} for further evaluation details.

The clear margin of \texttt{claude-4-opus} indicates that, for our task setup,
higher model–completeness and more coherent reasoning translate directly into more
actionable hydrological insights.  Conversely, despite producing the strongest
raw simulation summaries, \texttt{gpt-4o} lags on clarity, underscoring the need
for balanced optimisation across all evaluation axes.

\subsection{Selected Case Studies}
\label{subsec:case-study}
\paragraph{LLM-Vision-driven outlet gauge selection.}
The outlet–selection agent \(\mathcal{A}_{\mathrm{OS}}\) receives two georeferenced rasters—the flow-accumulation map and the DEM—together with vector layers of candidate gauges and their attributes (elevation, drainage area).  

\begin{figure}[htbp]
  \centering
  \includegraphics[width=1.0\linewidth]{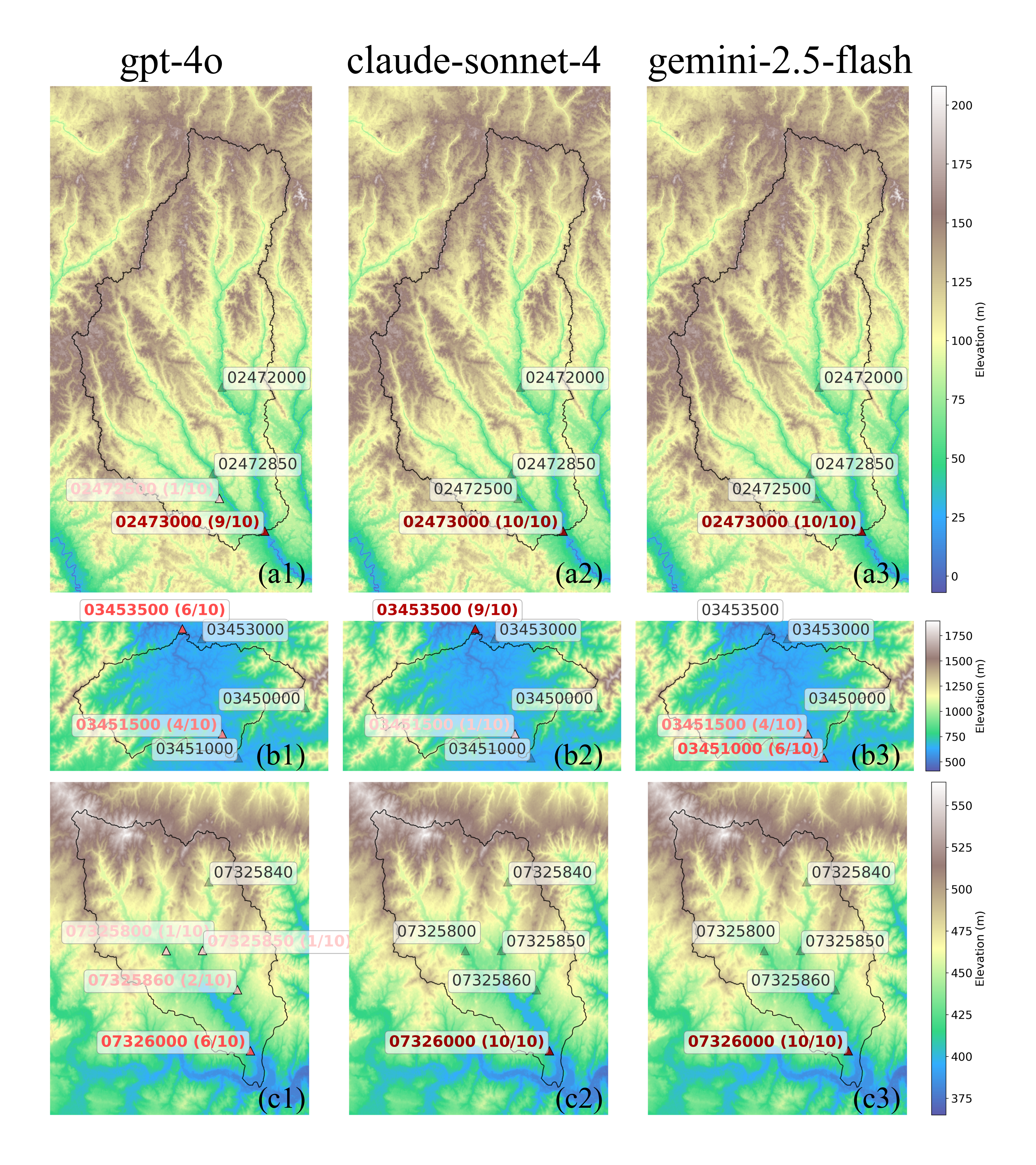}
  \caption{Gauge-selection frequency across three basins (rows a–c) and three LLMs (columns: GPT-4o, Claude-Sonnet-4, Gemini-2.5-Flash). Each basin–LLM combination was run ten times; gauge labels are shaded from black (never selected) to deep red (selected in all 10 runs).}
  \label{fig:outlet_selection}
\end{figure}

At inference time a VLM (GPT-4o, Claude-Sonnet-4, or Gemini-2.5-Flash) is prompted with these layers plus a short ordered rule set: (i) respect any user-specified gauge; (ii) disregard stations situated on or below reservoirs/lakes; (iii) favour the lowest-elevation gauge on the basin perimeter; (iv) prefer larger drainage areas and higher flow-accumulation values; and (v) break remaining ties with data-record quality.  All gauges inside the watershed, along with those lying just downstream of the polygon, are considered so that official outlets positioned slightly outside the boundary are not overlooked.

Figure~\ref{fig:outlet_selection} summarises the agent’s behaviour across three contrasting basins.  In the simple, single-outlet catchment (row~a) every LLM converges on the same gauge in almost every trial.  The more dendritic basin in row~b exposes subtle differences: GPT-4o and Claude-Sonnet-4 nearly always pick the hydrologically dominant tributary, whereas Gemini-2.5-Flash splits its choices between two interior stations, reflecting ambiguity in topographic cues.  Row~c highlights the importance of contextual constraints—when a major reservoir sits just upstream of the nominal outlet, all models occasionally retain the regulated gauge unless the prompt explicitly flags reservoir positions.


\begin{figure*}[t]
  \centering
  \includegraphics[width=0.9\linewidth]{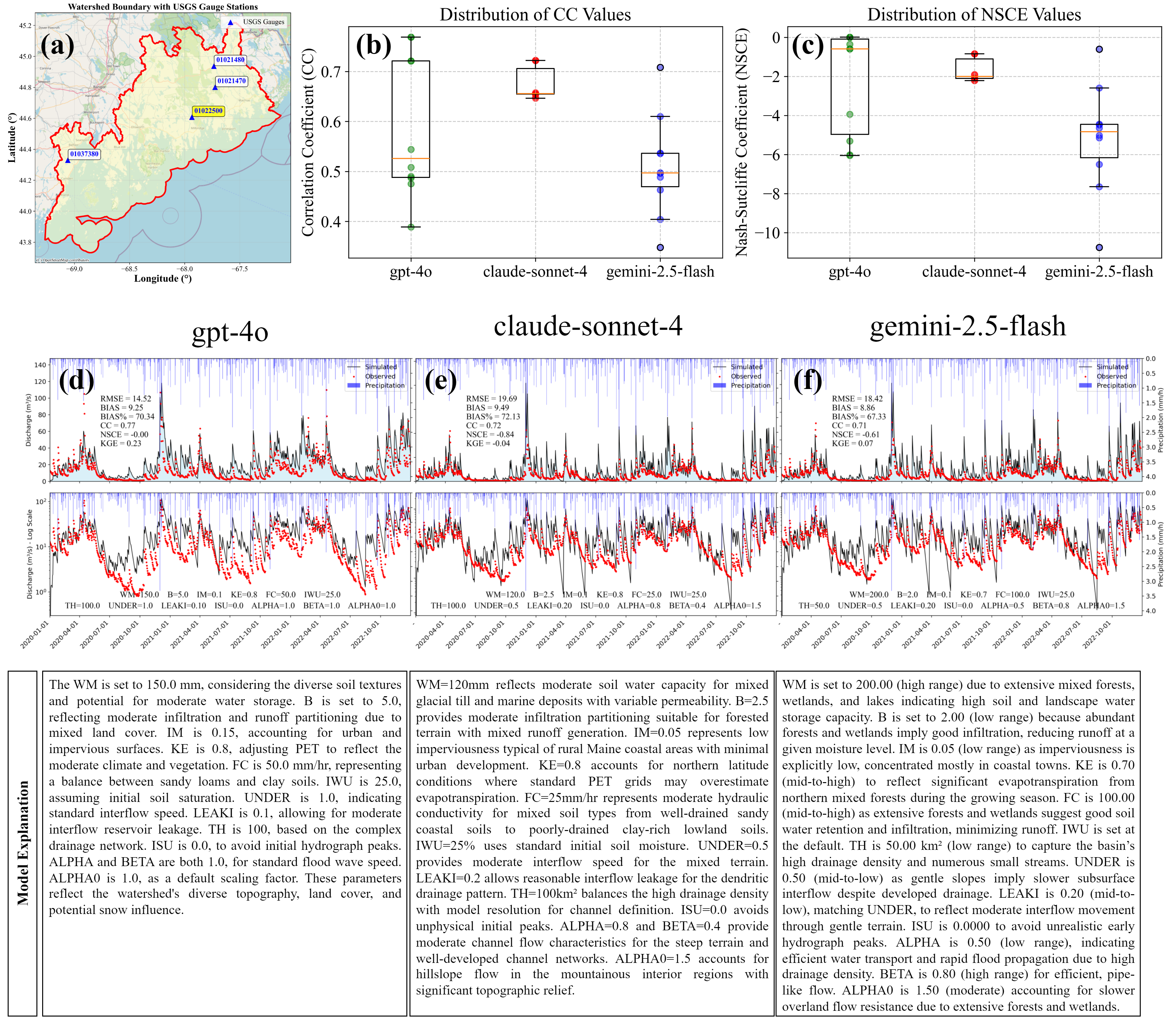}
  \caption{Performance summary for the Maine Coastal Basin.  
(a) Basin boundary and gauge locations.  
(b) Box-plot of CC from ten parameter-initialization runs for each LLM (GPT-4o, Claude-Sonnet-4, Gemini-2.5-Flash).  
(c) Corresponding NSCE box-plots.  
(d–f) Best-performing hydrographs for each LLM, annotated with their error metrics and calibrated CREST parameter sets.  
The lower text blocks provide the LLM-generated rationale for the chosen parameter values.}
  \label{fig:parameter_initial}
\end{figure*}

\paragraph{LLM-based first-guess parameterisation.}
Reliable calibration begins with a defensible \emph{first guess}, yet selecting a plausible vector of hydrologic model parameters normally demands years of field intuition and many trial–error cycles.  
AQUAH tackles this bottleneck with the agent \(\mathcal{A}_{\mathrm{PI}}\), which blends retrieval-augmented generation and vision reasoning.  
The agent consults CREST manuals(ingested as PDFs and web pages) to learn each parameter’s physical role and admissible range, while simultaneously analysing basin rasters—DEM, FAM, DDM, and a land-cover basemap—to infer slope, drainage density, soil moisture capacity, and impervious fraction.  Guided by this fused knowledge, the VLM proposes a basin-specific start vector (e.g.\ \(WM\), \(B\), \(KE\), see Appendix \ref{app:crest} for more information), runs CREST once, and logs the resulting skill scores.

Figure~\ref{fig:parameter_initial}b–c displays the distributions of CC and NSCE obtained from ten independent initialisations per LLM, while panels~(d–f) show each model’s best hydrograph and its associated parameter set.  The text blocks underneath capture the LLM-generated rationale—for example, boosting \(WM\) in forest-dominated headwaters or lowering \(B\) over urban sub-catchments.  Across the test basin the agent’s proposals consistently land within physically reasonable bounds and, in several cases, achieve near-calibrated performance on the very first run.  These results indicate that modern vision-LLMs already possess a rudimentary grasp of hydrologic parameter semantics, substantially shortening the path from “cold start” to productive calibration.

\section{Discussion}
\label{sec:discussion}
\paragraph{Gauge–Outlet Selection (Fig.~\ref{fig:outlet_selection}).}
Row~(a) represents a straightforward basin; all three LLMs consistently identify the true outlet, confirming that the rule set is sufficient for simple landscapes.  
Row~(b) introduces competing tributaries.  The correct outlet is gauge \texttt{03453500}; GPT-4o selects it in 6/10 trials, Claude-Sonnet-4 achieves 9/10, whereas Gemini-2.5-Flash never resolves the ambiguity.  
Row~(c) tests reservoir awareness: gauge \texttt{07326000} lies immediately downstream of a dam and should be rejected.  All models struggle—most runs still choose the regulated site—although GPT-4o avoids it in 40 \% of trials.  
These results highlight that discerning human regulation from natural flow remains challenging for current mainstream LLMs.  Incorporating additional Earth-observation cues (e.g.\ reservoir masks) or employing stronger reasoning models with tailored prompts may mitigate this limitation.

\paragraph{Parameter‐Initialisation Performance~(Fig.~\ref{fig:parameter_initial}).}
Across ten independent initialisations, the three LLMs exhibit distinct variance patterns.  
GPT-4o delivers the single highest CC/NSCE score but shows the widest spread, indicating strong stochasticity between runs.  
Claude-Sonnet-4 is the most consistent: its box-plots are narrow and uniformly positive, making it the most reliable performer for this basin despite not achieving the absolute best score.  
Gemini-2.5-Flash also displays high run-to-run variability but, unlike GPT-4o, its median skill is noticeably lower, leading to overall weaker performance.
In sum, GPT-4o can produce outstanding results but requires multiple attempts; Claude offers dependable, high-quality starts.

Although none of the first-guess parameter sets is fully optimal, each agent delivers values of the correct order of magnitude—an outcome far superior to ad-hoc, manual guessing and crucial for a successful first run.  Looking forward, the CREST outputs can be fed back to the \textit{ParamInit Agent}, enabling RAG-guided, step-wise adjustments that iteratively refine parameters.  Such a feedback loop would provide directionally consistent, interpretable calibration without exhaustive trial-and-error.
\paragraph{Limitation.}  
Our study reused one prompt template—originally tuned for the OpenAI API—across all language‑model back ends, which, while convenient for benchmarking, may not fully leverage each model’s unique formatting or capabilities. The prototype also depends on publicly hosted data and inference services (e.g., USGS, MRMS, and a cloud LLM), so regional access limits or temporary outages could reduce functionality. Future work will investigate model‑specific prompt tuning and local or cached data to ease these constraints.

\section{Conclusion}
\label{sec:Conclusion}

We present \textbf{AQUAH}, the first end-to-end hydrologic-simulation \emph{agent} that translates free-form language requests into physically consistent model runs and publication-ready reports. By coupling a large language model with vision modules for DEM reading, gauge detection, and parameter inference, AQUAH automates the full workflow---from data acquisition through CREST simulation to visualization---requiring neither domain expertise nor coding. Benchmarks across several mainstream LLMs show AQUAH delivers decision-grade outputs with low entry barriers.

Beyond hydrology, the modular design illustrates how \emph{LLM\,+\,CV} synergies can spawn specialised agents for other simulation-driven sciences across the globe.  We argue that building task-aware toolchains around foundation models will become a core paradigm for next-generation platforms, enabling rapid, democratised access to complex numerical engines across Earth-system, engineering, biomedical, environmental, and climate domains.

\section*{Acknowledgments}
\label{sec:Acknowledgments}
We gratefully acknowledge the anonymous hydrologic experts who evaluated our agentic modeling framework and provided valuable feedback.
This research was supported by NASA Award No.22‑IDS22‑0122 / 80NSSC24K0351, entitled “Predicting the Impact of Contemporary Climate Extremes on Nitrogen Flux along the Land‑to‑Ocean Continuum: Integrated Remote Sensing and Modeling Applied to the Mississippi River Basin.”
{
    \small
    \bibliographystyle{ieeenat_fullname}
    \bibliography{main}
}

\clearpage
\onecolumn
\appendix

\tcbset{
  promptbox/.style={
    colback=blue!5!white,
    colframe=blue!75!black,
    fonttitle=\bfseries,
    fontupper=\small
  }
}

\section{Experimental Setup}
\label{app:setup}

\paragraph{Implementation framework.}
All agents are orchestrated with \texttt{crewAI}~v0.75.0%
\footnote{\url{https://github.com/crewAIInc/crewAI}}, which provides the task queue, tool interface, and inter-agent messaging used throughout AQUAH.

\paragraph{ParamInitializer workflow.}
For illustration we focus on the \emph{ParamInitializer Agent}, whose logic is divided between two Python functions.  
\texttt{describe\_basin\_for\_crest()} prompts a vision-enabled LLM to summarise basin physiography from DEM, flow-accumulation, drainage-direction rasters, and a locator map;  
\texttt{estimate\_crest\_args()} then launches a CrewAI agent that mines PDF manuals and websites to propose a physically plausible CREST parameter vector.  
A provider-agnostic wrapper converts images to the base-64 or \texttt{PIL.Image} formats required by OpenAI, Anthropic, or Gemini APIs; oversized payloads are iteratively down-scaled and JPEG-compressed to satisfy the strictest quota (5 MB for Claude).

\paragraph{Large-language models.}
Five mainstream models are queried via their native endpoints: GPT-4o (\texttt{gpt-4o}), Claude-4 Sonnet (\texttt{claude-4-sonnet-20250514}), GPT-o1 (\texttt{o1}), Claude-4 Opus (\texttt{claude-4-opus-20250514}), and Gemini-2.5 Flash (\texttt{gemini-2.5-flash-preview-05-20}).  
Text-only prompts use a deterministic temperature of~0, whereas vision prompts use~0.3.

\paragraph{Earth-observation data.}
Input layers are fetched on demand from public repositories: HydroSHEDS 90 m DEM, flow-accumulation, and drainage-direction rasters (\url{https://hydrosheds.org/}); USGS 3DEP high-resolution DEMs (\url{https://apps.nationalmap.gov/downloader/}); MRMS precipitation archives (\url{https://mtarchive.geol.iastate.edu/}); FEWS-NET potential-evapotranspiration grids (\url{https://earlywarning.usgs.gov/fews/product/81}); and USGS NWIS discharge records (\url{https://waterdata.usgs.gov/nwis}).  
All layers are clipped to the basin polygon produced by the \textsc{ContextParser} agent and re-projected to a common grid before model execution.

\section{CREST}
\label{app:crest}
\paragraph{EF5/CREST model description.}
The EF5/CREST (Coupled Routing and Excess STorage) hydrologic modelling framework—originating from the University of Oklahoma in collaboration with NASA—combines distributed water-balance calculations with kinematic-wave routing to deliver rapid, spatially explicit flood simulations.  Over the past decade it has evolved into a versatile research and operational tool: CREST-iMAP couples hydrologic and hydraulic components for real-time inundation mapping \citep{li2021crest}; continental-scale calibration and validation have demonstrated robust skill across the CONUS domain \citep{chen2023conus}; the framework has been leveraged to diagnose forcing uncertainties such as the impact of IMERG precipitation upgrades on streamflow prediction \citep{zhu2024has}; and a recent synthesis highlights continued advances and emerging applications across global flood forecasting, drought assessment, and land–surface interaction studies \citep{li2023decadal}.  These studies underscore the model family’s breadth and its suitability for the automated, agent-driven workflows pursued in AQUAH.

\begin{figure}[htbp]
  \centering
  \includegraphics[width=\linewidth]{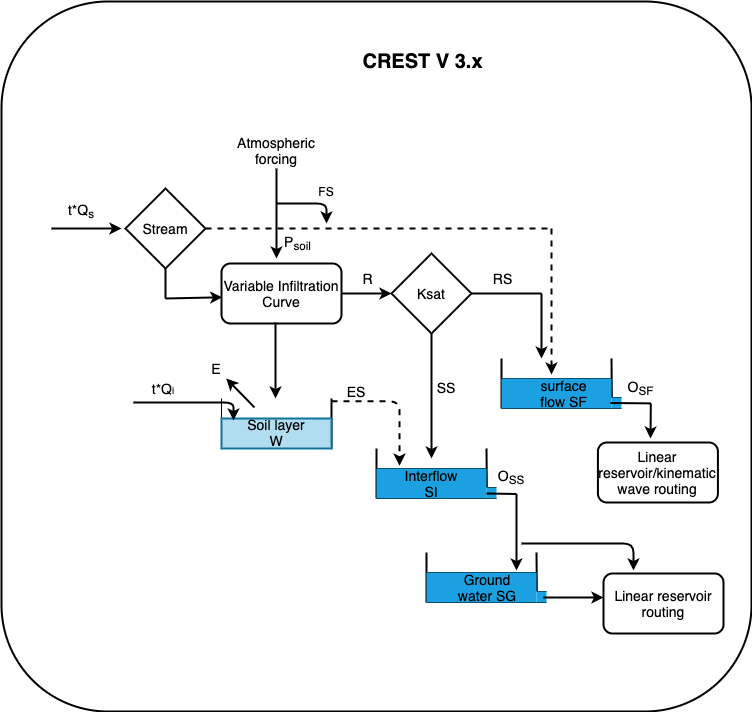}
    \caption{A schematic of the hydrologic processes represented by the latest EF5/CREST model}

  \label{fig:crest-model}
\end{figure}

\paragraph{EF5/CREST Parameter Cheat-Sheet.}

The EF5/CREST hydrologic model framework separates calibration parameters into two broad blocks: (i) \emph{runoff generation} governed by the CREST/Water-Balance scheme and (ii) \emph{kinematic-wave routing} \citep{flamig2020ensemble, li2023decadal}.  Tables \ref{tab:crest} and \ref{tab:kw} list the key parameters, their recommended search ranges, and the qualitative hydrologic response when each value increases.  This compact sheet is intended as a quick reference for modellers when setting up automatic or manual calibration routines.

\begin{table}[ht]
\centering
\caption{CREST / Water-Balance parameters}
\label{tab:crest}
\begin{tabular}{@{}l l c p{6.8cm}@{}}
\toprule
\textbf{Parameter} & \textbf{Meaning} & \textbf{Range} & \textbf{Effect when value increases}\\
\midrule
WM   & Maximum soil-water storage capacity (mm) & 5–250 & More storage $\Rightarrow$ \textbf{less} direct runoff.\\
B    & Infiltration curve exponent              & 0.1–20 & Steeper curve $\Rightarrow$ \textbf{more} surface runoff.\\
IM   & Fraction of impervious area          & 0.01–0.50 & Larger imperviousness $\Rightarrow$ \textbf{more} runoff.\\
KE   & PET utilisation / evapotranspiration coefficient & 0.001–1.0 & Higher ET loss $\Rightarrow$ \textbf{less} runoff.\\
FC   & Saturated hydraulic conductivity proxy (mm h\(^{-1}\)) & 0–150 & Faster infiltration $\Rightarrow$ \textbf{less} runoff.\\
IWU  & Initial soil-water content (mm)          & 0–25 & Wetter initial state $\Rightarrow$ \textbf{higher} early runoff.\\
\bottomrule
\end{tabular}
\end{table}

\begin{table}[ht]
\centering
\caption{Kinematic-wave routing parameters}
\label{tab:kw}
\begin{tabular}{@{}l l c p{6.8cm}@{}}
\toprule
\textbf{Parameter} & \textbf{Meaning} & \textbf{Range} & \textbf{Effect when value increases}\\
\midrule
TH      & Drainage-area threshold (km\(^2\)) & 30–300 & Smaller threshold $\Rightarrow$ finer channel network.\\
UNDER   & Interflow velocity multiplier (m s\(^{-1}\)) & 0.0001–3.0 & Larger velocity $\Rightarrow$ \textbf{quicker} runoff response.\\
LEAKI   & Leakage factor from interflow layer & 0.01–1.0 & Higher leakage $\Rightarrow$ faster hydrograph rise.\\
ISU     & Initial subsurface storage unit  & 0–\(1\times10^{-5}\) & Non-zero may cause spurious early peak; keep near zero.\\
ALPHA   & Muskingum–Cunge $\alpha$ for channel cells & 0.01–3.0 & Larger value \textbf{slows} flood-wave translation.\\
BETA    & Muskingum–Cunge $\beta$ for channel cells  & 0.01–1.0 & Bigger $\beta$ likewise slows and attenuates wave.\\
ALPHA0  & $\alpha$ for overland/non-channel cells & 0.01–5.0 & Controls overland flow speed; $\beta$ fixed at 0.6.\\
\bottomrule
\end{tabular}
\end{table}
\section{Evaluation Criteria}
\label{app:eval}


The quality of each AQUAH‐generated simulation is assessed through a
two-tier protocol that combines \emph{objective statistical metrics} and
\emph{human expert review}.  The former quantify the numerical agreement
between simulated and observed discharge, while the latter capture
practitioner-oriented aspects such as interpretability and report
readability.

\paragraph{Objective Verification Metrics}
Following established hydrological practice, five complementary statistics are evaluated over the full period (see Table~\ref{tab:verification_metrics}). These are: the \textit{Nash–Sutcliffe efficiency} (NSE, $-\infty$–1, ideal 1), which summarises overall predictive skill; the \textit{Kling–Gupta efficiency} (KGE, ideal 1) that balances correlation, bias and variability; the \textit{Pearson correlation coefficient} (CC, ideal 1); the \textit{root mean square error} (RMSE), where lower values indicate smaller deviations; and the \textit{relative bias} (BIAS), whose optimum is 0. Together they diagnose both the accuracy and reliability of the CREST simulations across all flow regimes.

\begin{table*}[htbp]
\centering
\caption{Verification metrics used in this study.  
$Q_{\text{obs}}^{t}$ ($Q_{\text{sim}}^{t}$) is the observed (simulated) discharge at time step~$t$;  
$\overline{Q}_{\text{obs}}$ and $\overline{Q}_{\text{sim}}$ are their respective means;  
$\mu$ and $\sigma$ are the mean and standard deviation; $T$ is the total number of time steps.  
\textit{CC} – Pearson correlation coefficient, \textit{BIAS} – relative bias, \textit{RMSE} – root mean square error,  
\textit{NSE} – Nash--Sutcliffe efficiency, \textit{KGE} – Kling--Gupta efficiency with  
$\alpha=\sigma_{\text{sim}}/\sigma_{\text{obs}}$ and $\beta=\mu_{\text{sim}}/\mu_{\text{obs}}$.  
The last column gives each metric’s theoretical range and its perfect value (in parentheses).}
\label{tab:verification_metrics}
\renewcommand{\arraystretch}{1.25}
\begin{tabular}{@{}l l c@{}}
\toprule
\textbf{Metric (abbr.)} & \textbf{Equation} & \textbf{Range (perfect)} \\
\midrule
Nash--Sutcliffe efficiency (NSE) &
$\displaystyle
NSE = 1-
\frac{\sum_{t=1}^{T}(Q_{\text{obs}}^{t}-Q_{\text{sim}}^{t})^{2}}
     {\sum_{t=1}^{T}(Q_{\text{obs}}^{t}-\overline{Q}_{\text{obs}})^{2}}$
& $(-\infty,\,1]$ (1) \\[1.0em]

Relative bias (BIAS) &
$\displaystyle
BIAS = \frac{1}{T}\sum_{t=1}^{T}\!
       \bigl(Q_{\text{sim}}^{t}-Q_{\text{obs}}^{t}\bigr)$
& $(-\infty,\,\infty)$ (0) \\[1.0em]

Root mean square error (RMSE) &
$\displaystyle
RMSE = \sqrt{\frac{1}{T}\sum_{t=1}^{T}
       \bigl(Q_{\text{sim}}^{t}-Q_{\text{obs}}^{t}\bigr)^{2}}$
& $[0,\,\infty)$ (0) \\[1.0em]

Correlation coefficient (CC) &
$\displaystyle
CC = \frac{\sum_{t=1}^{T}
       (Q_{\text{sim}}^{t}-\overline{Q}_{\text{sim}})
       (Q_{\text{obs}}^{t}-\overline{Q}_{\text{obs}})}
     {\sqrt{\sum_{t=1}^{T}
       (Q_{\text{sim}}^{t}-\overline{Q}_{\text{sim}})^{2}}\,
      \sqrt{\sum_{t=1}^{T}
       (Q_{\text{obs}}^{t}-\overline{Q}_{\text{obs}})^{2}}}$
& $[-1,\,1]$ (1) \\[1.2em]

Kling--Gupta efficiency (KGE) &
$\displaystyle
\begin{aligned}
KGE &= 1-\sqrt{(CC-1)^{2}+(\alpha-1)^{2}+(\beta-1)^{2}} \\[0.3em]
\end{aligned}$
& $(-\infty,\,1]$ (1) \\
\bottomrule
\end{tabular}
\end{table*}

\begin{figure}[htbp]
  \centering
  \includegraphics[width=0.95\linewidth]{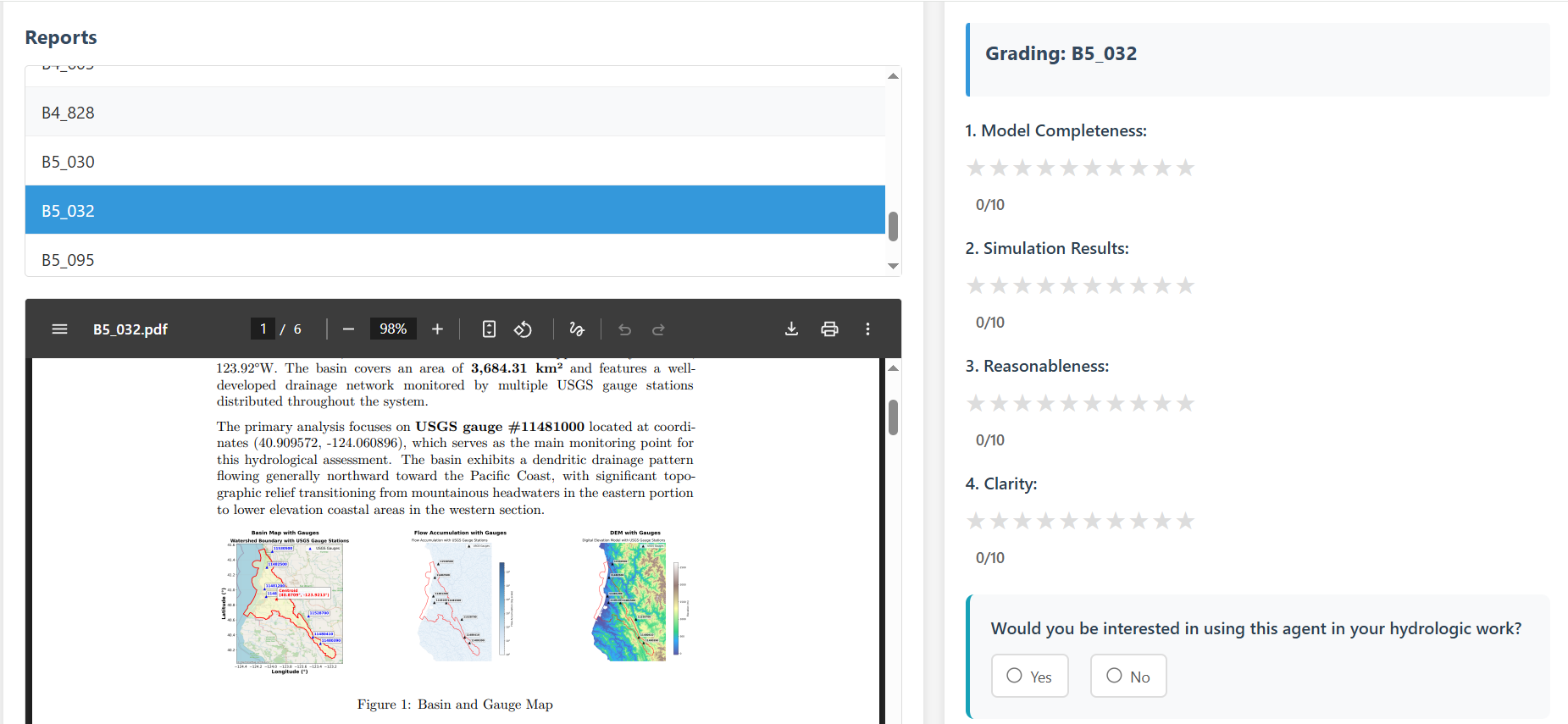}
  \caption{Human‐grading interface used in this study.  Experts (and an
  LLM co-evaluator) assign 0–10 star scores on four axes—Model
  Completeness, Simulation Results, Reasonableness, and
  Clarity—and record whether they would adopt the agent in professional
  hydrologic work.}
  \label{fig:grading_ui}
\end{figure}

\paragraph{Final Report Evaluation}
Beyond the objective metrics, every report is first uploaded to the latest \texttt{o3} large-language model for automated grading and then independently assessed—under blinded conditions—by a team of professional hydrologists, both parties applying the same four-axis rubric. \textit{Model Completeness} gauges the suitability of data sources, openness of parameter disclosure, and overall workflow transparency; \textit{Simulation Results} reflects the fidelity of hydrographs and accompanying statistics, including treatment of uncertainties; \textit{Reasonableness} judges the physical plausibility of parameter choices, underlying assumptions, and recommended next steps; and \textit{Clarity} measures readability, logical flow, figure and table quality, and adherence to scientific-writing norms. Each axis is scored on an integer 0–10 scale by the expert panel and the LLM; the two values are averaged to obtain the axis score, and the unweighted mean across the four axes yields an overall quality index (see the UI mock-up in Fig.~\ref{fig:grading_ui}).

\paragraph{Example Analysis}

\begin{figure}[htbp]
  \centering
  \includegraphics[width=0.95\linewidth]{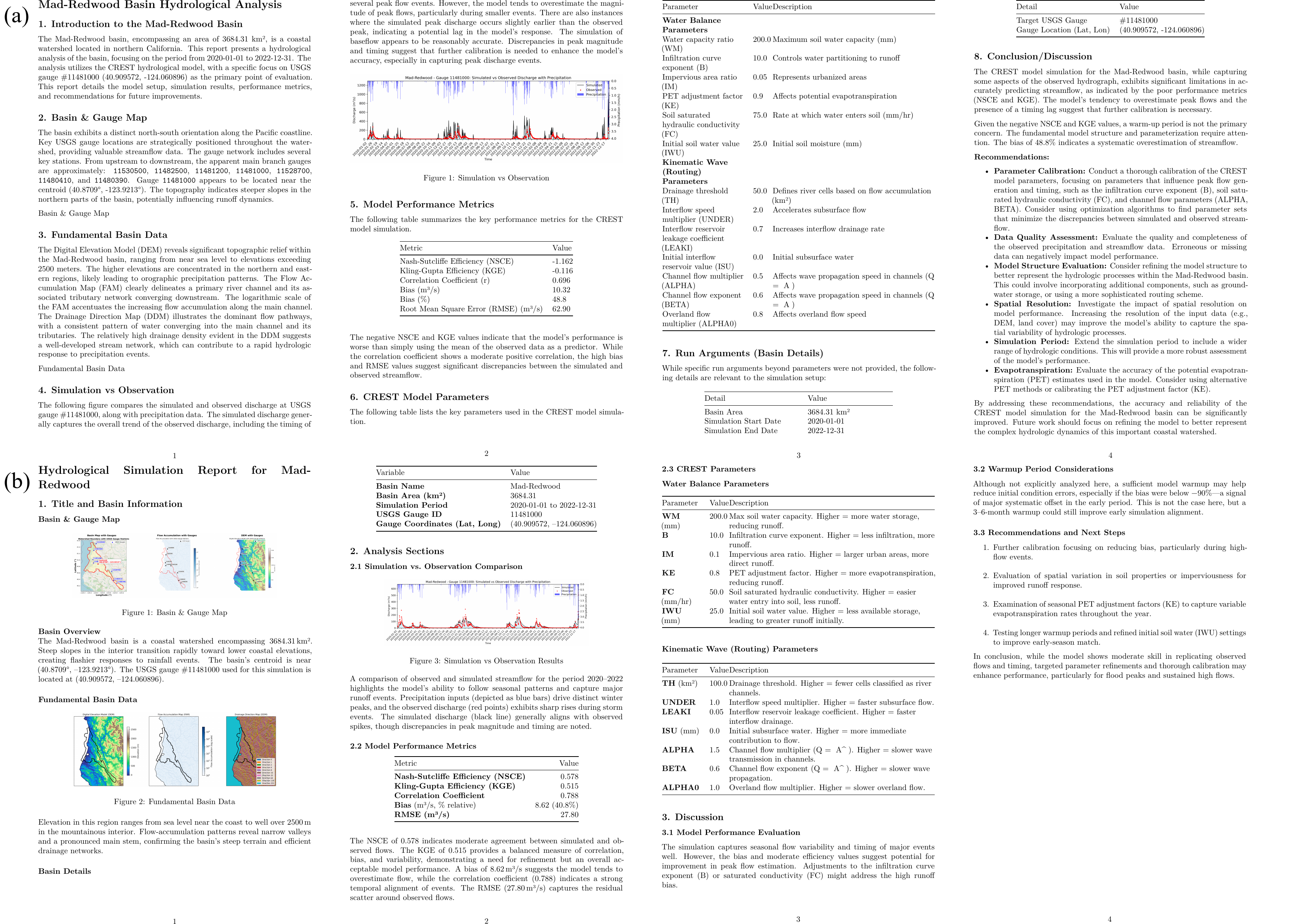}
  \caption[Grading example for two agent-generated reports]{Side-by-side grading example for two hydrological reports generated by different LLM agents. Panel (a) shows the report B5\_030.pdf produced by \texttt{gemini-2.5-flash}, while panel (b) displays B5\_223.pdf from \texttt{gpt-o1}. Both were created from the same prompt, “\textit{I want to simulate the streamflow of the Mad–Redwood basin from 2020 to 2022}.” The table underneath presents the averaged human + LLM scores on the four-axis rubric. Owing to missing figures, B5\_030 lags in \emph{Model Completeness}; its poorer NSE also lowers the \emph{Simulation Results} score. In contrast, B5\_223 achieves notably higher marks across all axes, leading to a superior overall quality index.}
  \label{fig:report_review_example}
\end{figure}

Figure~\ref{fig:report_review_example} contrasts two reports generated from the identical prompt “\textit{I want to simulate the streamflow of the Mad–Redwood basin from 2020 to 2022}.”  Panel (a) shows \emph{B5\_030.pdf}, produced by the \texttt{gemini-2.5-flash} agent, while panel (b) shows \emph{B5\_223.pdf} from \texttt{gpt-o1}.  Although both agents follow the same workflow, their outputs diverge noticeably: \texttt{B5\_030} omits several key figures, lowering its \emph{Model Completeness} score, and its poor NSE drags down the \emph{Simulation Results}.  In contrast, \texttt{B5\_223} includes all requisite graphics and attains a substantially better NSE (0.578), which, together with clearer recommendations, yields higher marks across all four grading axes and a superior overall index.  This example underscores how agent choice can strongly influence both the technical fidelity and presentation quality of first-pass hydrologic simulations.

\end{document}